\documentclass[letterpaper, 10 pt, conference]{ieeeconf}  

\IEEEoverridecommandlockouts                              

\usepackage{hyperref}
\usepackage[pdftex]{graphicx}
\usepackage{algorithm}
\usepackage{algorithmic}
\usepackage{amsmath}
\usepackage{booktabs} 
\usepackage{marvosym}
\usepackage{subfigure}
\usepackage{siunitx}

	\usepackage{fancyhdr}
	
\title{\LARGE \bf
RIDM: Reinforced Inverse Dynamics Modeling for Learning from a Single Observed Demonstration*
}

\author{Brahma S. Pavse$^{\text{\Cross}, 1}$, Faraz Torabi$^{\text{\Cross}, 1}$, Josiah Hanna$^{2}$, Garrett Warnell$^{3}$, and Peter Stone$^{4}$
\thanks{*This work has taken place in the Learning Agents Research
	Group (LARG) at UT Austin.  LARG research is supported in part by NSF
	(CPS-1739964, IIS-1724157, NRI-1925082), ONR (N00014-18-2243), FLI
	(RFP2-000), ARO (W911NF-19-2-0333), DARPA, Lockheed Martin, GM, and
	Bosch.  Peter Stone serves as the Executive Director of Sony AI
	America and receives financial compensation for this work.  The terms
	of this arrangement have been reviewed and approved by the University
	of Texas at Austin in accordance with its policy on objectivity in
	research.}
\thanks{$^{\text{\Cross}}$First Author and Second Author had equal contribution}
\thanks{$^{1}$First Author and Second Author are with Computer Science Department, University of Texas at Austin, USA
	{\tt\footnotesize brahmasp@utexas.edu, faraztrb@cs.utexas.edu}}%
\thanks{$^{2} $Third Author is with School of Informatics, University of Edinburgh, Scotland. To be joining the Computer Sciences Department, University of Wisconsin-Madison, USA.
	{\tt\footnotesize josiah.hanna@ed.ac.uk}}%
\thanks{$^{3} $Fourth Author is with Army Research Laboratory, USA
	{\tt\footnotesize garrett.a.warnell.civ@mail.mil}}%
\thanks{$^{4} $Fifth Author is with Computer Science Department, University of Texas at Austin and Sony AI, USA. 
	{\tt\footnotesize pstone@cs.utexas.edu}}
}

\begin{document}

\thispagestyle{fancy}

\maketitle
\thispagestyle{fancy}

\begin{abstract}

Augmenting reinforcement learning with imitation learning is often hailed as a method by which to improve upon learning from scratch.
However, most existing methods for integrating these two techniques are subject to several strong assumptions---chief among them that information about demonstrator actions is available.
In this paper, we investigate the extent to which this assumption is necessary by introducing and evaluating {\em reinforced inverse dynamics modeling} (RIDM), a novel paradigm for combining imitation from observation (IfO) and reinforcement learning with no dependence on demonstrator action information.
Moreover, RIDM requires only a single demonstration trajectory and is able to operate directly on raw (unaugmented) state features.
We find experimentally that RIDM performs favorably compared to a baseline approach for several tasks in simulation as well as for tasks on a real UR5 robot arm. Experiment videos can be found at \href{https://sites.google.com/view/ridm-reinforced-inverse-dynami}{https://sites.google.com/view/ridm-reinforced-inverse-dynami}.

\end{abstract}

\section{INTRODUCTION}

Two of the most prevalent paradigms for behavior learning in artificial agents are imitation learning (IL) \cite{NIPS1996_1224, Argall:2009:SRL:1523530.1524008} and reinforcement learning (RL) \cite{Sutton98reinforcementlearning}.
Agents that use IL receive a strong training signal in the form of an expert demonstration, but because their goal is to imitate, their task performance is typically bounded above by that of the expert.
Agents using RL, on the other hand, can theoretically learn behaviors that are optimal with respect to a predefined task reward, but often have difficulty doing so due to practical challenges such as large state spaces and sparse reward functions.
Because of the relative advantages and disadvantages of each of these paradigms, it is natural to investigate whether one can integrate them in order to get the best of both methods.

While combining IL and RL has been explored to a certain extent in the literature  \cite{Taylor:2011:IRL:2031678.2031705, lakshminarayanan2016reinforcement,  DBLP:journals/corr/abs-1802-09564}, several important issues remain.
%
Most importantly, these techniques require access to the internal control signals used by a demonstrator in order to be able to leverage the demonstration information \cite{NIPS2016_6391, DBLP:journals/corr/abs-1802-09564, DBLP:journals/corr/HesterVPLSPSDOA17}.
This requirement makes it difficult to obtain useful demonstrations since it necessitates a high level of internal access to the demonstration platform, preventing, e.g., the use of numerous, easily-accessible video demonstrations available on websites like YouTube.
A second limitation of many existing techniques is the requirement for {\em many} expert demonstrations \cite{bojarski2016end}, which makes obtaining sufficient demonstration data difficult in that it requires a high level of access to expert demonstrators.
Finally, existing methods typically assume that they have access to {\em task-specific} state features during the learning process that can be used to make learning easier \cite{NIPS2016_6391, IJCAI2018-torabi, torabi2019generative}.
Task-specific state features are ones that somehow augment the agent's natural (or {\em raw} state information using additional domain knowledge---like the distance to certain important subgoals---designed to make reward function representation easier (see, e.g., Figure \ref{fig:aug_states}).
While providing this domain knowledge may be fairly easy for a specific task, it will, in general, need to be specified anew for each new task encountered and therefore represents a practical impediment to using existing methods.


\begin{figure}[]
	\centering{\includegraphics[scale=0.5]{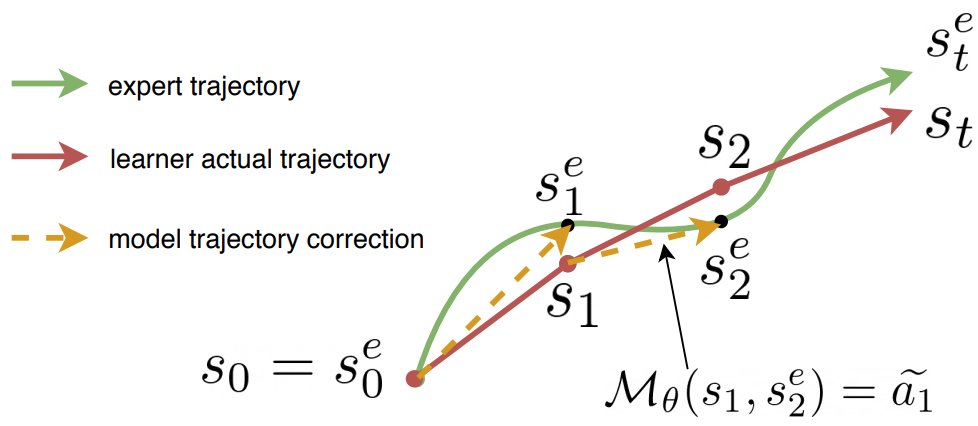}}
	\vspace{-6mm}
	\caption{RIDM applies a task-specific inverse dynamics model,  $\mathcal{M}_{\theta}$, on the learner's current state, $s_t$, to the expert's next state, $s_{t+1}^e$, such that the sequence of executed actions, $\{\widetilde{a_t}\}$, maximizes the cumulative reward from the environment. At each time step, the agent uses the expert's next state, $s_{t+1}^e$ (black dot), as the set point for $\mathcal{M}_\theta$. However, it actually reaches $s_{t+1}$ (red dot) instead---which is typically {\em not} the set point---since RIDM optimizes $\mathcal{M}_\theta$ to {\em maximize environment task reward} instead of minimizing trajectory-tracking error.}
	\label{fig:ridm_schematic}
	\vspace{-3mm}
\end{figure} 

In this paper, we propose a new technique for integrating IL and RL called {\em reinforced inverse dynamics modeling (RIDM)} that bypasses the issues identified above.
RIDM leverages recent ideas from model-based imitation from observation (IfO) to enable integrated imitation and reinforcement learning from a {\em single, action-free} demonstration consisting of only {\em raw states}.
Moreover, RIDM represents a new paradigm for combining IL and RL in that the agent's behavior is based on following a fixed demonstration trajectory using a parameterized
task-specific inverse dynamics model (IDM) (see, Figure \ref{fig:ridm_schematic}). A task-specific IDM is one that maps state-transitions to actions for a specific task only and
and may not generalize to other tasks that have different reward functions.
While RIDM requires the demonstration trajectory during execution, its overall objective is {\em not} to imitate, but rather to maximize the external environment reward.
RIDM accomplishes this by using RL to tune the IDM that {\em attempts} to follow the fixed demonstration such that the resulting behavior leads to the highest environmental reward.
Formulating the overall RL problem in this way allows RIDM to diverge from the demonstration if doing so will lead to higher task reward, which is helpful when the demonstration is sub-optimal.
To the best of our knowledge, we are the first to introduce an algorithm that combines IfO and RL.

To evaluate our algorithm, we establish a baseline algorithm by modifying a state-of-the-art IfO algorithm to incorporate an external reward signal.
We hypothesize that RIDM will be able to outperform this baseline 
in the problem of interest where a few, raw-state demonstration is provided.
We perform several quantitative experiments focused on both simulated and real robot control tasks, and find that RIDM's unique, model-driven approach results in high-quality behavior trajectories that lead to better performance than the baseline.

\section{Related Work}
\label{related_work}
This section provides a broad outline of research related to our work. 
The section is organized as follows. 
Section \ref{related_ifo_control} details the most related works on imitation from observation and reinforcement learning.
Section \ref{related_rl_and_il} discusses efforts in integrating reinforcement learning and imitation learning.

\subsection{Imitation from Observation and Reinforcement Learning}
\label{related_ifo_control}
The focus of imitation from observation (IfO) \cite{DBLP:journals/corr/LiuGAL17, torabi2019recent} is to learn a policy that results in 
similar behavior as the expert demonstration with state-only
demonstrations. There are broadly two approaches: (1) model-based
and (2) model-free. In our work, we are focused on
model-based approaches. 



Many model-based IfO algorithms use an inverse dynamics model, i.e., a mapping from state-transitions to actions. 
The most related work to ours may be the work of Nair et al.\cite{DBLP:journals/corr/NairCAIAML17}, who show the learner a single demonstration of an expert performing some task with the intention of the learner replicating the task.
Their algorithm allows the learner to undergo self-supervision by collecting states and actions, which are then used to train a neural network inverse dynamics model.
The learned model is then applied on the expert demonstration to infer the  expert actions. 
The actions are then executed to replicate the demonstrated behavior. 
Another method is behavioral cloning from observation (BCO)
\cite{IJCAI2018-torabi}, which consists of
two phases. The first phase trains an inverse dynamics model in
a self-supervised fashion, and applies the learned
model on the expert demonstration(s) to infer the expert
actions. The second phase involves training a policy by behavioral cloning (BC) \cite{Pomerleau1991}, which maps the expert states to the inferred actions. BCO, however, does not factor in the environment reward
to train the inverse dynamics model or policy in either of the phases. 

Iterative learning control (ILC) \cite{bristow2006survey} is an older trajectory tracking approach which operates in a repetitive manner to improve its tracking precision. The methods developed in ILC, often use PID controllers and attempt to optimize the PID gains so that the agent follows the reference trajectory more accurately. Our work differs from ILC in that our objective is not accurate trajectory tracking, but rather to maximize the available environment reward, and we use RL to achieve that objective
The benefit of our work is that if the demonstration is sub-optimal, the final learned behavior could potentially outperform the demonstrator's.

The focus of reinforcement learning is to train agent to learn a task in an environment by maximizing some notion of cumulative reward. In our work, we are focused on using black-box optimization methods. Some of the most related works are as follows. Hwangbo et al. \cite{hwangbo2014rock} propose a method, ROCK*, for tuning a PD controller that
performs favorably to CMA-ES on their experiments.
Calandra et al. \cite{calandra2016bayesian} use Bayesian Optimization to tune a
state machine for robotic locomotion. They also test their
method on a linear controller (which corresponds to a PD
controller if the states contain positions and velocities).
Neuman-Brosig et al. \cite{neumann2019data} apply Bayesian optimization for
learning the parameters for active disturbance rejection
control.  Leonetti et al. \cite{leonetti2012combining} use controlled random search
to tune a linear controller. Black box optimization for
controller tuning has also been applied in several
undergrad theses and reports \cite{marco2015gaussian}. Our work differs from the mentioned past work in this area in that we integrate IfO and RL which potentially helps with constraining the amount of exploration required for learning the behavior.

\subsection{Integrating Reinforcement Learning and Imitation Learning}
\label{related_rl_and_il}

Another area of research related to our work is dealing with
the case in which an expert demonstration may be sub-optimal. One way to address this issue
is by combining reinforcement learning and imitation learning.

There has been significant effort to combine reinforcement learning and imitation learning. For example, 
Taylor et al. \cite{Taylor:2011:IRL:2031678.2031705} introduced
Human-Agent Transfer, an algorithm that uses a human
demonstration to build a base policy, which is further
refined using reinforcement learning on a robot soccer
domain.
Lakshminarayanan et al.\cite{lakshminarayanan2016reinforcement} uses a hybrid 
formulation of reward and expert state-action
information in the replay buffer when training deep Q-network (DQN) to speed-up
the training procedure. Hosu et al.\cite{DBLP:journals/corr/HosuR16}
use deep RL to learn an Atari game but they use human checkpoint 
replays as starting points during 
the learning process instead of re-starting the game at the
end of the episode. Subramanian et al.\cite{Subramanian:2016:EDI:2936924.2936990} and Nair et al.\cite{DBLP:journals/corr/abs-1709-10089} use IL information to
alleviate the exploration process in RL. Hester et al.\cite{DBLP:journals/corr/HesterVPLSPSDOA17} pre-train a deep
neural network by optimizing a loss that includes a temporal difference
(TD) loss as well as supervised learning loss with the expert
actions. Zhu et al. \cite{DBLP:journals/corr/abs-1802-09564} optimize
a linear combination of the imitation reward outputted by generative
adversarial imitation learning (GAIL) \cite{NIPS2016_6391} and
the task reward. These
works assume that the learner has access to the expert's actions.

%
Our work is distinct from all these works in that we focus
on the integration of reinforcement learning and imitation from observation
where we only have access to expert state trajectories -- \textit{not} the expert actions. 




\section{Preliminaries}
\label{preliminaries}
We begin by reviewing and establishing notation for reinforcement learning, imitation learning, and imitation from observation.


\subsection{Reinforcement Learning (RL)}
\label{prelim_rl}

We model agents interacting in an environment as a Markov decision process (MDP).
A MDP is denoted by the tuple $M = \langle S, A, T, R\rangle$, where $S$ is the state space of the agent, $A$ is the action space of the agent, $T$ defines the environment transition function that gives the probability of the agent moving from one state to another given that the agent took a particular action (i.e., $T: S \times A \times S \rightarrow [0,1]$), and $R$ is the scalar-valued reward function that dictates the reward received by the agent when moving from one state to another via a particular action.
In the context of the MDP framework, the reinforcement learning problem is that of optimizing the agent's behavior so as to find a control policy, $\pi^\star: S\rightarrow A$, that the agent can use to maximize the total cumulative reward it receives.

\subsection{Imitation Learning (IL)}
\label{prelim_lfd}


In contrast to reinforcement learning, imitation learning involves a learner seeking to mimic the behavior of an expert demonstrator rather than maximizing an external reward signal.
We denote demonstrations as $D^e = \{(s_t^e, a_t^e)\}$,  where $s^e_t$ denotes the state of the expert at a given time index $t$, and $a^e_t$ denotes the action taken by the expert at that time.
Given one or many such demonstrations, the goal of IL is to learn a control policy $\pi$ that the learning (imitating) agent can use to produce behavior similar to that of the expert.
%

IL in the absence of expert action information, i.e., when $D^e = \{s_t^e\}$, is called imitation from observation (IfO).
The IfO problem is that of learning the same imitation policy $\pi$ as in IL, but without access to this action information.
The learner is shown only the \emph{states} of the expert.
In this case, most IL methods no longer apply, and we must find new strategies.
We might try, for example, to infer the expert actions $\{a_t^e\}$ to get
$\{\widetilde{a_t^e}\}$ for each state $\{s_t^e\}$, and therefore approximate $\widetilde{D^e} = \{(s_t^e, \widetilde{a_t^e})\}$ and then apply conventional IL methods as done by Torabi et al. \cite{IJCAI2018-torabi}.
In this work, we study the problem of integrating IfO with RL.



\section{Reinforced Inverse Dynamics Modeling}
\label{method}

We now introduce reinforced inverse dynamics modeling (RIDM) -- a new method for integrating IfO and RL.
RIDM learns a strategy by which an agent can select actions $\{\widetilde{a_t}\}$ that allow it to achieve a high level of task performance when it has available a single, state-only expert demonstration, $D^e = \{s_t^e\}$.

RIDM does so by learning and using a task-sepcific inverse dynamics model (IDM), $\mathcal{M}_\theta$, that infers which action to take at any given time instant based on both the agent's current state and a desired next state, the {\em set point}.
Under RIDM, the agent's actions are computed as $\widetilde{a_t} = \mathcal{M}_\theta(s_t,s^e_{t+1})$, where $s_t$ is the learner's current state and $s^e_{t+1}$ is the state of the expert at the next time instant.
The goal of RIDM is to find an optimal $\theta$ such that the generated action sequence maximizes the cumulative reward from the environment, $R_{env}(\mathcal{M}_{\theta})$.
That is, while RIDM selects which actions to take by using the expert's state sequence as a sequence of set points, it evaluates its policy in relation to the {\em environmental reward} as opposed to, e.g., the trajectory-tracking error.
Note that it may actually be {\em desirable} for the induced state sequence to differ from that of the expert's if doing so allows for higher environment reward.
Figure \ref{fig:ridm_schematic} depicts this process.
To the best of our knowledge, using such a scheme to perform integrated IfO and RL is unique in the literature.


RIDM consists of two phases.
The goal of the first phase is to initialize the inverse dynamics model.
This phase can either be done by selecting $\theta$ at random, or -- if a known policy is available to the learner -- by having the agent generate its own set of state-action-next-state triples and using supervised learning to fit $\theta$ to those triples.
In the second phase, RIDM alternates between generating agent behavior according to $\theta$ and the expert demonstration, and optimizing $\theta$ in response to the amount of environment reward obtained by the generated behavior.
During this phase, the learner uses the demonstration to guide the agent's behavior (i.e., imitation), but uses the observed environment reward to adjust $\theta$ such that actions leading to high rewards are generated (i.e., reinforcement).
The goal of this two-phase procedure is to find the optimal policy in terms of total task reward (which may outperform the expert) by using the expert demonstration as a guide.
The pseudo-code for RIDM is given in Algorithm \ref{method_code}, and each phase is described in more detail below. \footnote{Even though RIDM is described as a method that requires both environment rewards and state-only demonstrations, the algorithm can be used even if the reward is not available for instance by defining the reward as the negative of the distance between the demonstrated state and the imitator's state at each time step.}

\begin{algorithm}
	\caption{RIDM}
	\begin{algorithmic}[1] 
		\label{method_code}
		\REQUIRE Single, state-only demonstration $D^e := \{s_t^e\}$
		\label{algo:idminit}
		\IF{$\pi^{pre}$ available} \label{algo:pretrainstart}
		\STATE Generate $D^{pre} := \{(s^{pre}_t, a^{pre}_t)\}$ using $\pi^{pre}$
		\STATE Initialize $\theta$ as the solution to (\ref{pre_fitness})
		\ELSE
		\STATE Initialize $\theta$ uniformly at random
		\ENDIF \label{algo:pretrainend}
		\WHILE{$\theta$ not converged} \label{algo:ridmstart}
		\FOR{$t = 0:|D^e| - 1$}
		\STATE $\widetilde{a_{t}} := \mathcal{M}_\theta(s_t, s_{t + 1}^{e})$
		\STATE Execute $\widetilde{a_{t}}$ and record $s_{t+1}$ and reward $r_t$
		\ENDFOR
		\STATE{Compute cumulative episode reward $R_{env} = \sum_t r_t$}
		\STATE{Update $\theta$ by solving (\ref{main_fitness})}
		\ENDWHILE \label{algo:ridmend}
		\RETURN $\theta^*$
	\end{algorithmic}
\end{algorithm}

\subsection{Inverse Dynamics Model Pre-training}
\label{idm_pre_train}

During RIDM's optional first phase, an initial value for $\theta$ is sought.
This initialization is accomplished either through the use of data collected by the learner using a pre-defined exploration policy or, if such a policy is not available, by selecting the parameter value at random.
We allow for RIDM to take advantage of an available exploration policy so that it can achieve a reasonable level of task performance, which is likely to get us into a good basin of attraction within the optimization landscape.




In the case where an exploration policy $\pi^{pre}$ is 
available (e.g. if a slow-walk policy is available and
we want the agent to learn a fast walk), RIDM computes an initial value for $\theta$ as follows.
First, the learner executes $\pi^{pre}$ in the environment and records the resulting experience as a trajectory of length $T$ that we denote as $D^{pre} = \{(s^{pre}_t, a^{pre}_t, s^{pre}_{t+1})\}$.
The initial value for $\theta$ is then computed by solving the following supervised learning problem:
\begin{align}
\label{pre_fitness}
\theta^* = \arg\max \Bigg( -\frac{1}{T} \sum_{t=1}^{T} \sum_{n=1}^{N} \frac{\lvert 
	\mathcal{M}_\theta(s_t^{pre}, s_{t + 1}^{pre})_n - a_{tn}^{pre}\rvert}{\max{(a_{n}^{pre})} - \min{(a_{n}^{pre}})}\Bigg) \; ,
\end{align}
where $N$ is the dimensionality of the action space, $a_{tn}^{pre}$ denotes the scalar value of the $n^\text{th}$ component of the action vector $a^{pre}_t$, and $\max(a_n^{pre})$ denotes the maximum value of $a_{tn}^{pre}$ across all $t$.
Above, notice that the goal of the optimization problem is to select $\theta$ such that $\mathcal{M}_\theta(s^{pre}_t, s_{t + 1}^{pre})_n$ is a good approximation of the true action value $a_{tn}^{pre}$.
We adopt the particular loss given above because we found that it worked well in practice.
It is able to effectively trade off short-term errors in order to optimize the differences across a full trajectory, and the normalization term ensures greater accuracy for actions which vary over a smaller range.
RIDM solves (\ref{pre_fitness}) using a blackbox optimization technique (e.g., CMA-ES\cite{Hansen:2003:RTC:772374.772376}). Note, however, that this pre-training
phase is optional, and only possible when RIDM has
access to an exploration policy that generates a behavior
that is qualitatively similar to the desired end behavior.

\subsection{Inverse Dynamics Model Reinforcement}
\label{idm_reinforcement}

RIDM's required second phase seeks to iteratively update the inverse dynamics model parameters in response to the environment return.
The process executed here is illustrated in Figure \ref{fig:ridm_schematic}, where one can see that RIDM uses the expert's demonstration as a behavior template in the sense that the expert's state trajectory is used as a sequence of set points to guide behavior.

The iterative updates to $\theta$ are computed as follows.
First, the learner uses $\mathcal{M}_\theta$ and the expert demonstration to generate a trajectory of experience.
It does so by, when in state $s_t$ at time step $t$, executing action $\widetilde{a_t} = \mathcal{M}_\theta(s_t, s_{t+1}^e)$, which results in a transition to state $s_{t+1}$ and the observation of reward $r_t$.
After this trajectory has been generated, the learner computes the cumulative environment reward $R_{env}(D^e \; ; \; \theta) = \sum_t r_t$, which is dependent on both the (fixed) expert demonstration data $D^e$ and the (tunable) model parameters $\theta$.
In a given iteration, $i$, an update to $\theta$ is computed as the solution to:
\begin{equation}
\label{main_fitness}
\begin{aligned}
\theta_i = \arg\max R_{env}(D^e \; ; \; \theta_{i - 1}) \; .
\end{aligned}
\end{equation}
It is important to note that, here, expert's actions are \textit{unknown}. While $R_{env}(D^e \; ; \; \theta)$ is used to reinforce the learning of the inverse dynamics model parameters, the learner is always guided by the same, fixed, state-only expert demonstration trajectory.

For each iteration of the above procedure, RIDM solves (\ref{main_fitness}) again using a blackbox optimization technique (eg., CMA-ES\cite{Hansen:2003:RTC:772374.772376}
or Bayesian optimization\cite{pelikan1999boa}).




\section{Empirical Results}
\label{empirical_results}

We now empirically validate our hypothesis, i.e., that behaviors learned using RIDM will outperform those learned by the established baseline.
We focus on the case in which only a single, state-only demonstration is available to the agent and no task-specific state augmentation can be performed.
Our experiments are executed in multiple robot control domains: simulated tasks are carried out in the MuJoCo and SimSpark simulators, and several manipulation tasks are carried out on a UR5 robot arm. 
In the first set of experiments with the MuJoCo simulator, neural networks are used to model IDMs since such models have proven effective in these domains in the literature. A neural network IDM gets one (state,next-state) pair as input and outputs the Gaussian distribution parameters from which an action is to be sampled and executed by the agent.
In the rest of the experiments, i.e., the SimSpark simulator and physical tasks, PID controllers are used to model IDMs since, again, these have proven effective in these domains in the past. PID controllers are not exactly inverse dynamics models due to their differential and integral terms. However, they retain the essential characteristic that RIDM requires, i.e., that, given a current state and desired next state (or set point), they will generate an action that attempts to reach the set point.
In the training process, the next state (or set point) at each time step is fixed according to the demonstration, and RIDM uses reinforcement learning to tune the IDM parameters (either NN parameters or PID gains) such that the overall agent behavior can best maximize an external environment reward. RIDM uses CMA-ES\cite{Hansen:2003:RTC:772374.772376} in all the simulated
experiments, but uses Bayesian optimization\cite{pelikan1999boa} in the physical robot experiments to learn the inverse dynamics model due to its superior sample complexity.



This section is organized as follows.
First, we provide experimental motivation for studying RIDM in the single, state-only demonstration and raw state space case.
Next, we establish a reasonable IfO+RL baseline that can also operate in this regime, and we validate our hypothesis by comparing RIDM to this baseline.
In each of our evaluations, we scale the reported performance metrics such that a score of $0$ corresponds to the behavior of a random policy, and a score of $1$ corresponds to the behavior of the expert.
Note that, when the expert performance is sub-optimal, because we are combining imitation learning with reinforcement learning, it is reasonable to expect that the algorithm will outperform the expert on some tasks.
Finally, we conclude this section by reporting additional empirical results for applying RIDM to both simulated robot soccer skill learning and to learning to perform a behavior on a physical UR5 arm robot.

\subsection{Experimental Setup}

Prior research has established that the availability of demonstrator action information and many demonstration trajectories are critical for the success of existing imitation learning algorithms \cite{IJCAI2018-torabi, torabi2019generative}.
While RIDM is advantageous in that it does not depend on the availability of the above information, we have also claimed that it can operate directly on raw state information that has not been augmented using extra knowledge of the task which, of course, is typically unknown or difficult to obtain.
We now seek to experimentally motivate the need for overcoming this issue by showing the level of reliance on these augmented state spaces in many existing imitation and reinforcement learning algorithms.

\begin{figure}[]
	\centering{\includegraphics[scale=0.37]{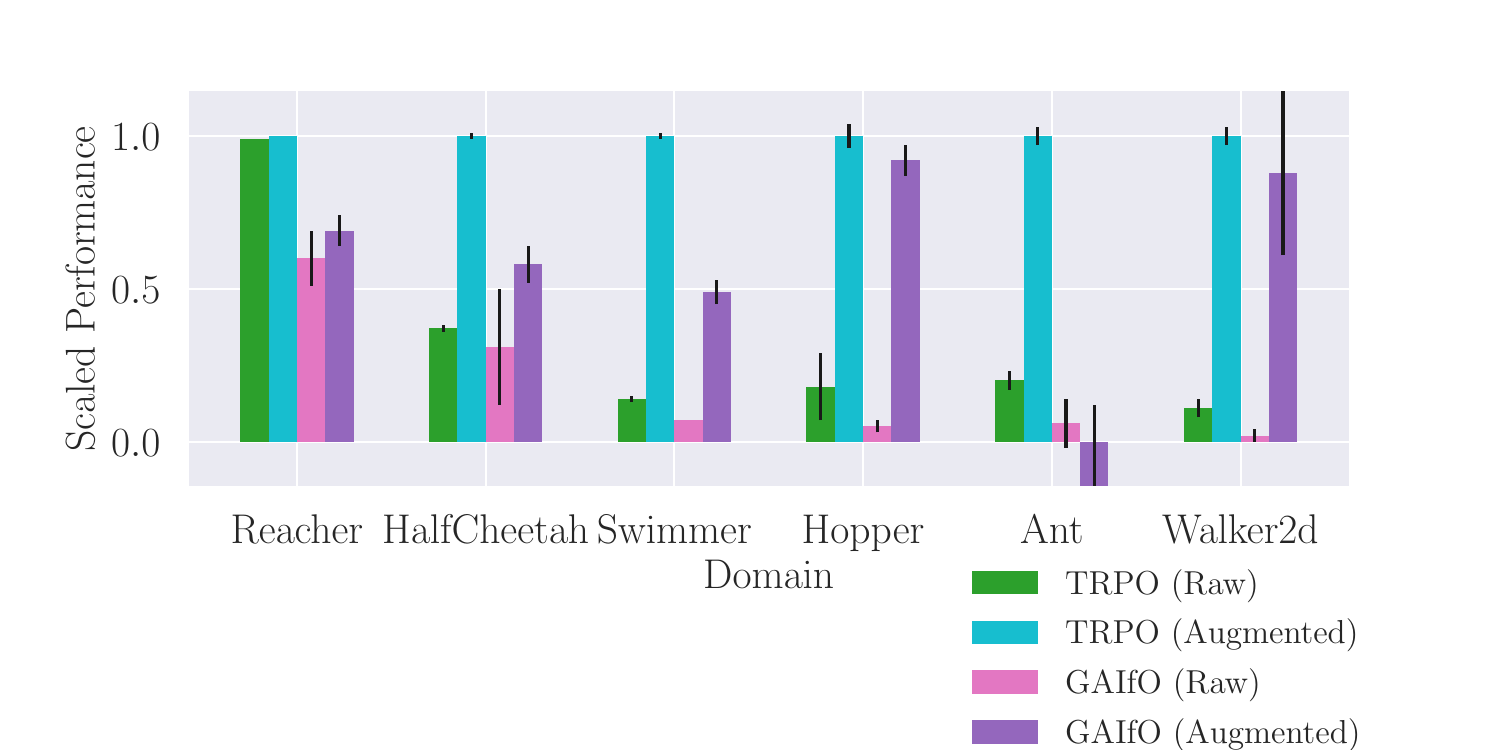}}
	\vspace{-6mm}
	\caption{Quantitative exhibition of the importance of an augmented state
		space for high performance on six MuJoCo domains for GAIfO  and TRPO. Mean and standard deviations are over $100$
		policy runs. Both methods
		use a neural network parameterized policy.}
	\label{fig:aug_states}
	\vspace{-3mm}
\end{figure}

In Figure \ref{fig:aug_states}, we compare the scaled performance of an imitation from observation algorithm (GAIfO\cite{torabi2019generative}) and a reinforcement learning algorithm (TRPO\cite{DBLP:journals/corr/SchulmanLMJA15}) when they are given access to an augmented state space vs. when they are exposed to only the raw state space on six tasks from the MuJoCo domain \cite{DBLP:conf/iros/TodorovET12} \footnote{The MuJoCo experiments use all the standard settings, e.g., the reward functions, the goals, and the augmented state spaces, as defined in the MuJoCo code base.}. Here, the raw state space refers to the list of joint angles, and the augmented state space also includes task-specific information. For instance, in the Hopper task, the augmented state space includes the agent's global position which is advantageous in that it is more highly correlated with the reward signal (see, e.g., \cite{torabi2019generative, torabi2019imitation, NIPS2016_6391, DBLP:journals/corr/SchulmanLMJA15, DBLP:journals/corr/SchulmanWDRK17}).
In this specific task, the agent's goal is that of controlling the limbs of a $2$D, one-legged robot such that it moves forward as fast as possible.
The task reward given per time step corresponds to the change in global position of the agent, and since this information appears in the augmented state information, both learning algorithms perform much better (except for GAIfO in the Ant domain, perhaps due to the very large ($111$ dimensional) augmented state space). This advantage can be seen in Figure \ref{fig:aug_states}, where the high reliance of  GAIfO\cite{torabi2019generative} and TRPO\cite{DBLP:journals/corr/SchulmanLMJA15} on the augmented state space is readily apparent.
Such augmentations are, in general, restrictive in that they need to be redefined for each new task.

Because we seek to remove the above restriction, we use only the joint angles as the raw state information in our experiments.
Many robots are comprised of joints, and therefore 
a joint-only representation is reasonably task-independent. The core results of our algorithm are exclusively concerned with dealing with the above case, i.e., single state-only demonstration consisting of joint angles.

\subsection{RIDM Applied to MuJoCo Simulation}
We now present our core results. This section
is organized as follows. Section \ref{baseline_gaifo_rl} proposes
a reasonable baseline to compare against our method. Section \ref{core_hypothesis_valid} presents the performance of our algorithm
against this baseline.

\subsubsection{Baseline: GAIfO+RL}
\label{baseline_gaifo_rl}
To the best of our knowledge, there is no method in the existing literature which can operate in the experimental setting of interest.
Therefore, in order to understand the effectiveness of RIDM, we first propose the natural combination of the best existing algorithms for the components of RIDM, namely IfO and RL as the baseline.

GAIfO+RL is based on the current state-of-the-art imitation from observation algorithm, GAIfO\cite{torabi2019generative}.
Starting from GAIfO, GAIfO+RL integrates RL by modifying the reward function used during the agent update step.
Instead of the reward function being determined solely by the discriminator as in GAIfO, GAIfO+RL integrates imitation and reinforcement learning by defining a new reward function that is a linear combination of the discriminator's output and the task reward \cite{DBLP:journals/corr/abs-1802-09564}.

\begin{figure}[]
	\centering{\includegraphics[scale=.34]{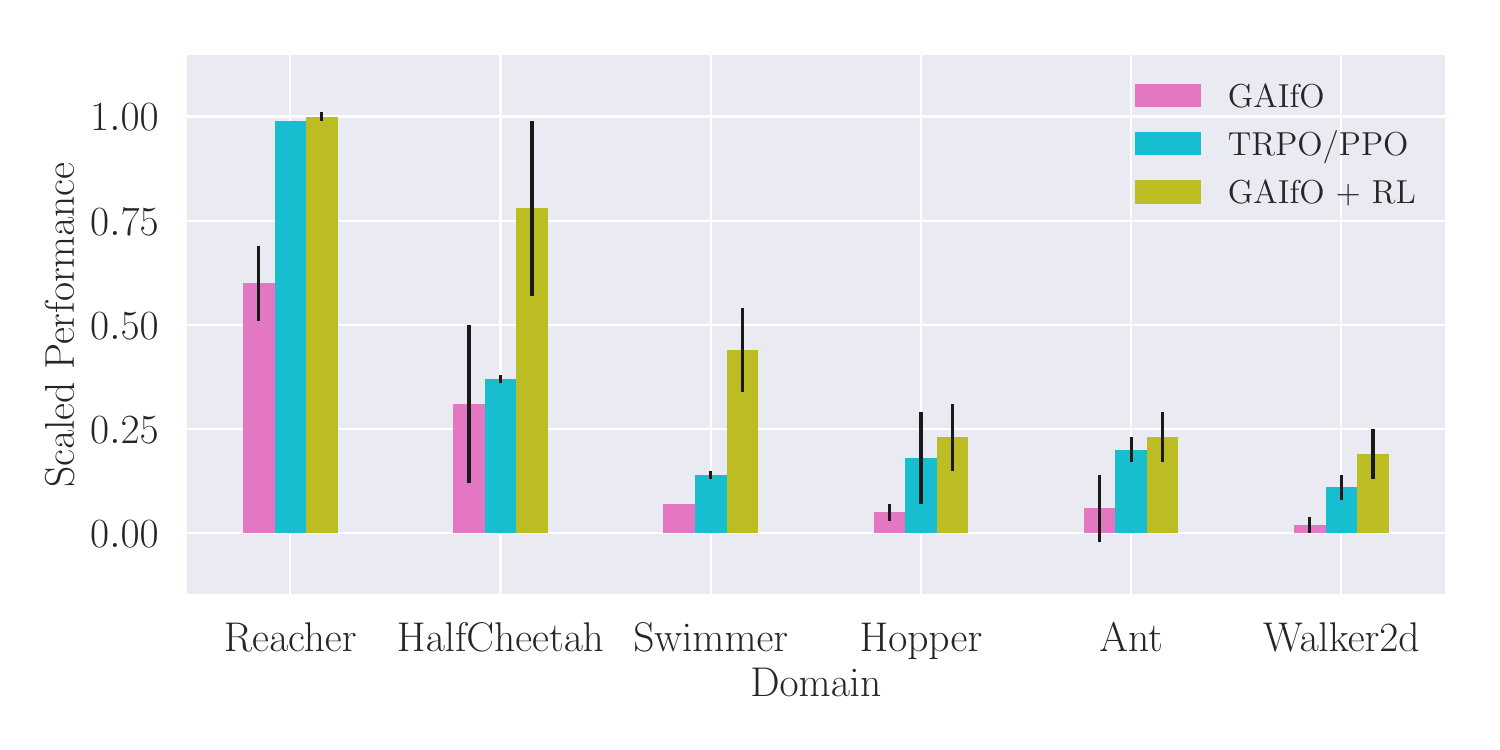}}
	\vspace{-8mm}
	\caption{Establishment of GAIfO+RL as a reasonable IfO+RL baseline to compare against RIDM. All methods use the same single state-only demonstration
		consisting of only raw states (exclusively of joint angles).
		Mean and standard deviations are over $100$
		policy runs. All methods
		use a neural network parameterized policy.}
	\label{fig:gaifo_rl_baseline}
	\vspace{-3mm}
\end{figure}

In Figure \ref{fig:gaifo_rl_baseline}, we establish that GAIfO+RL is a strong baseline by evaluating the performance of GAIfO alone, RL alone (TRPO/PPO), and GAIfO+RL.
All three methods operate in the raw (un-augmented) state space, and GAIfO and GAIfO+RL are also given access to a single, state-only demonstration.
While the performance of either pure IfO or pure RL alone is relatively poor, we can see that GAIfO+RL achieves significantly higher performance than its parts.
Moreover, GAIfO+RL operates in the same established regime and belongs to the same class of IfO+RL algorithms as RIDM, and therefore seems to be a reasonable imitation from observation + reinforcement learning algorithm.

\subsubsection{Hypothesis Validation}
\label{core_hypothesis_valid}

\begin{figure}[]
	\centering{\includegraphics[scale=.34]{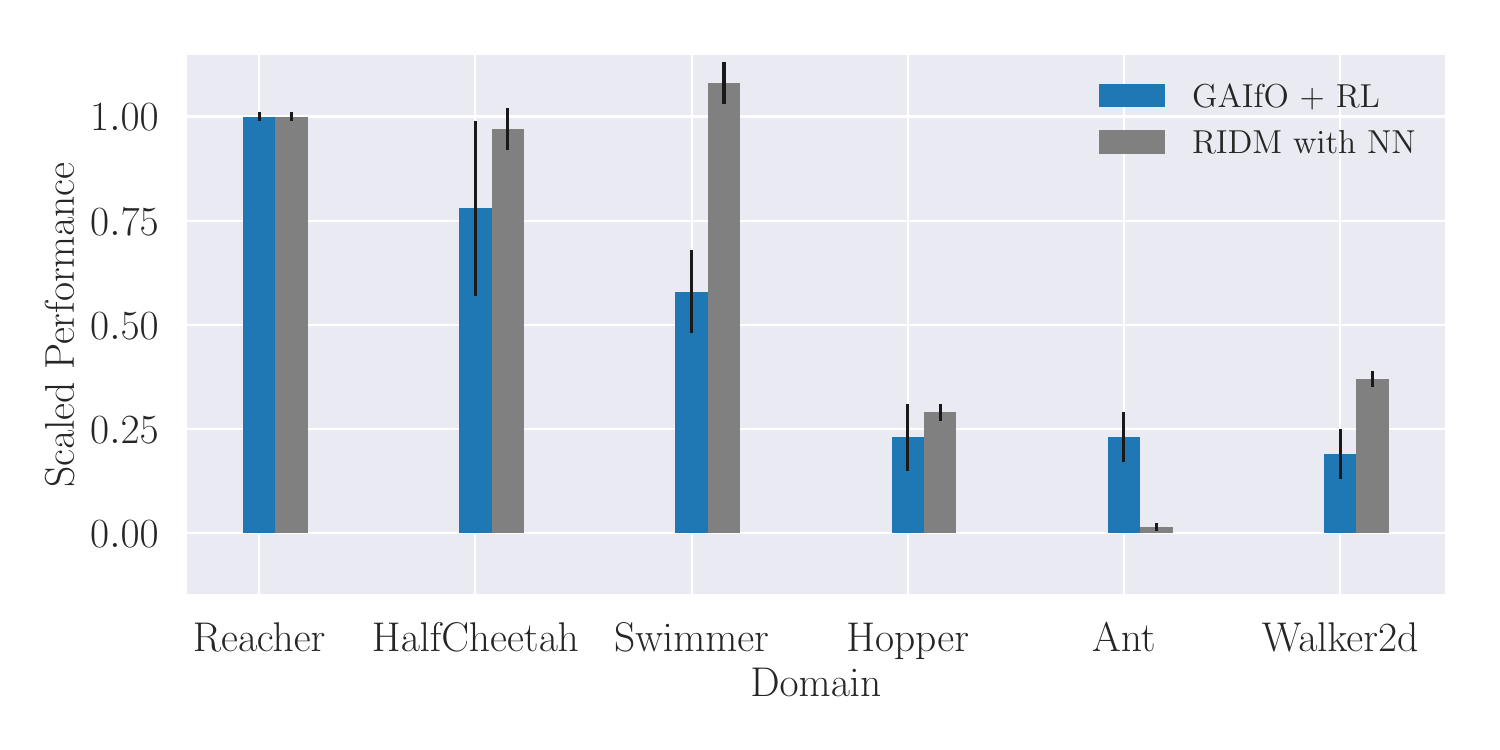}}
	\vspace{-8mm}
	\caption{Comparison of RIDM final performace against established baseline, GAIfO+RL, on the MuJoCo domain on the same single state-only demonstration
		consisting of only raw states (exclusively of joint angles).
		Mean and standard deviations for GAIfO+RL and RIDM are over $100$
		policy runs. GAIfO+RL uses a neural network parameterized policy.
		For each domain, in order
		of $x$-axis, the numbers of iterations required for RIDM are  $700$, $800$, $400$, $100$, $900$, and $1300$ and for GAIfO+RL are  $400$, $800$, $1000$, $1000$, $1200$, and $1500$}
	\label{fig:mujoco_nn}
	\vspace{-3mm}
\end{figure}

We conducted an experiment comparing
the scaled performance of RIDM against that of the GAIfO+RL baseline 
on six tasks from the MuJoCo domain. The experts are generated using TRPO/PPO with augmented states. However, the demonstrated trajectories only include the raw state information. 
Here, we first model the inverse dynamics model using a neural network and train the network to maximize the received reward while attempting to follow the expert trajectory. The results are presented in Figure \ref{fig:mujoco_nn}. It can be seen that RIDM outperforms GAIfO+RL in five of the domains. The only domain that the performance is worse than the baseline is the Ant domain. We speculate that the neural network IDM is not able to learn a meaningful model due to the complexity of the domain resulting from the large number of joints compared to each of the other domains. In Section \ref{add_results}, we show
that if RIDM uses a lower-dimensional parameterized IDM (e.g.
a PID controller), the performance of the learning agents is improved.

\begin{figure}[]
	\centering{\includegraphics[scale=.34]{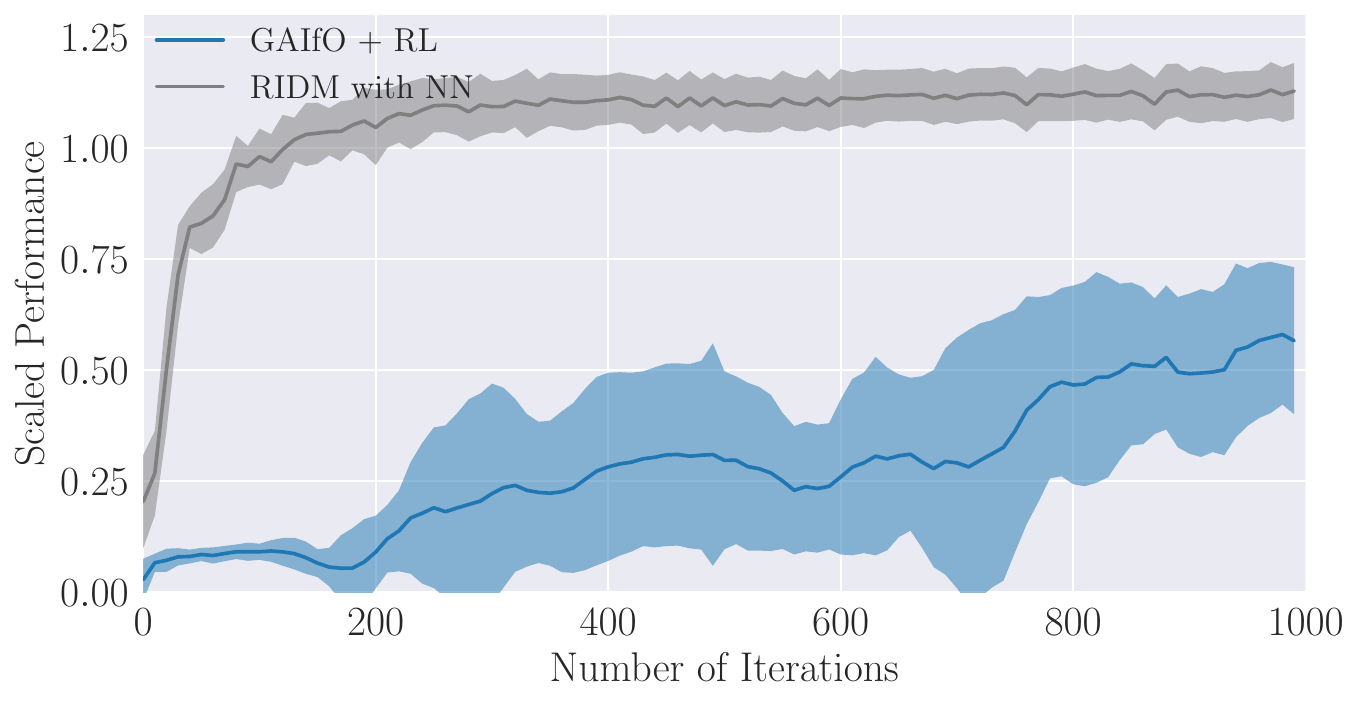}}
	\vspace{-1mm}
	\caption{Comparison of RIDM learning process against established baseline, GAIfO+RL, on the Swimmer domain on the same single state-only demonstration
		consisting of only raw states (exclusively of joint angles). Solid lines represent the mean return and shaded areas represent standard deviations over $10$ trials. While the shown graph is for Swimmer-v2, we observed the same qualitative trend on other domains as well.}
	\label{fig:process}
	\vspace{-2mm}
\end{figure}

%

\subsection{RIDM Applied to SimSpark RoboCup $3$D Simulation}
We now report the results of using RIDM to learn agent behaviors in the RoboCup 3D simulation environment, SimSpark\cite{Bdecker2008SimSparkC, 10.1007/978-3-662-44468-9_59}.
Specifically, our goal was to determine whether or not RIDM could imitate agent skills exhibited by the agents of other teams that participate in the RoboCup 3D simulation competition \cite{LNAI18-MacAlpine}. Since the
opponent's policies are unknown, we obtain the
demonstration by executing the teams' computer-readable but non-human-readable code in the environment.

In our experiments, we are interested in two tasks: (1) speed walking, and (2) long distance kick-offs. We collect demonstration data of two teams, FC Portugal (FCP) \cite{reis2017fc} and FUT-K \cite{iwanagafut}. 
RIDM pre-trained the model (see Section \ref{idm_pre_train}) using 
walk and kick exploration policies from our own team, UT Austin Villa\cite{LNAI18-MacAlpine}.
Here, we report results using only RIDM since it proved infeasible to evaluate 
GAIfO+RL in this domain due to the computational
time complexity. We found that RIDM performed best with
global PD gains common to all joints as the inverse dynamics model.

Below are the reward function details of each task:
\begin{itemize}
	\item Speed walking: Summation of
	distances (meters) travelled per time-step with a $-5$
	penalty for falling down.
	\item Long-distance kick-off: 
	\begin{align*}
	R_{kick} = (1 + x_{total}) \cdot \exp {\Big(\frac{-\theta^2}{180}\Big)} + x_{air} \cdot 100
	\end{align*}
	with a penalty $-5$ for bumping the ball, $-10$ for falling down,
	where $x_{total}$ is the $x$-axis distance traveled by the ball, $\theta$ is the angle between the ball's trajectory and the
	line between the agent and center of
	the goal, and $x_{air}$ is the $x$-axis distance for which the ball was traveling in
	the air. Distances are in meters, and
	$\theta$ is in degrees.
\end{itemize}
Since we defined these reward functions independently from the demonstrations, the demonstrations do not optimize the reward signals. The demonstrators are trained for the RoboCup task; their performances are sub-optimal with regards to our designed reward functions.

Tables \ref{robocup_walk_results} and \ref{robocup_kick_results} and  summarize our results.
We report both the performance of the expert and our agent.
We can see that, since RIDM takes advantage of both the reward functions and the demonstrations, it allows our agents to {\em outperform} the sub-optimal experts.

\begin{table}[]\centering
	\caption{RIDM vs. expert for speed walking.}
	\begin{tabular}{@{}lllll@{}} \toprule
		\label{robocup_walk_results}
		Expert & Agent & Speed (m/s)  & Reward  \\ \midrule
		FCP 
		
		& \textbf{RIDM (ours)} & \textbf{0.81} & \textbf{9.82} \\ 
		& Expert & 0.69 & 8.35 \\
		\midrule
		FUT-K  
		
		& \textbf{RIDM (ours)} & \textbf{0.89} & \textbf{10.70}\\ 
		& Expert & 0.70 & 8.47 \\ \bottomrule
		
	\end{tabular}
\end{table}

\begin{table}[]\centering
	\caption{RIDM vs. expert for long-distance kick offs.}
	\begin{tabular}{@{}llllll@{}} \toprule
		\label{robocup_kick_results}
		Expert & Agent  & $x_{air}$ (m) & $x_{total}$ (m) & Reward \\ \midrule
		FCP 
		& \textbf{RIDM (ours)} & \textbf{13.78} & \textbf{24.05}  & \textbf{1386.00} 
		
		\\ 
		& Expert & 8.00 & 17.00 & 808.00 \\ \midrule
		
		FUT-K 
		
		& \textbf{RIDM (ours)}  & \textbf{10.62} & \textbf{16.23}  & \textbf{1064.00} \\ 
		& Expert & 0.00 & 10.00  & 1.00 \\ \bottomrule
	\end{tabular}
\vspace{-3mm}
\end{table}

\subsection{RIDM Applied to a Physical UR5 Robot Arm}
We also used RIDM for behavior learning on a physical robot.
Specifically, we used a UR5, a 6-degree-of-freedom robotic arm.
We considered a reaching task in which the arm begins in a consistent, retracted position, then must move its end effector (i.e., the gripper at the end of the arm) to a target point in Cartesian space, and finally must stop moving once the end effector has reached the target point.
We trained the expert by iterating between iLQR \cite{tassa2012synthesis} and dynamics learning with a specified reward function.
We then executed this expert policy and recorded the resulting trajectories to create the demonstration data \cite{torabi2019sample}.

For the physical arm experiments, we skip the
pre-training phase for two reasons: 1) we did not
have access to a sub-optimal policy for each task,
and 2) for safety concerns, we did not want to use
a random exploration policy.
For RIDM's second phase, we used a reward function defined as the negative of the Euclidean distance of the end effector of the arm to the target point at each timestep. We used Bayesian optimization\cite{pelikan1999boa} as
the blackbox optimization algorithm to update the model parameters in response to the environment reward.
Bayesian optimization works by constructing a posterior distribution over the space of functions being optimized over.
Here, this distribution was represented using a Gaussian process over functions that map PID values to the episode returns.
As the training proceeds and more data is observed, Bayesian optimization techniques sharpen the posterior, resulting in more certainty as to which regions of the parameter space are worth exploring further with more trials and which are not.
For simpler optimization problems, Bayesian optimization is more sample efficient compared to CMA-ES and converges within a few iterations.

Table \ref{tb:ur5-results} represents the results of our experiments on the UR5, where we compare RIDM to a baseline behavior generated by using the demonstration state sequence as set points for the platform's pre-defined, hard-coded PID controller parameterization. The reported numbers are the averages and standard deviations of episode returns over five separate experiments all of which are reaching tasks with different target points.
Table \ref{tb:ur5-results} shows 
that while
RIDM outperforms the original PID, it is worse than the expert. The reason is that
the expert is optimal with regards to the designed reward function.

\begin{table}[]\centering
	\caption{RIDM vs. Original PID controller vs. Expert.}
	\begin{tabular}{@{}lllll@{}} \toprule
		\label{tb:ur5-results}
		Agent & Reaching & Pushing & Pouring  \\ \midrule
		RIDM (ours)  & -11.94 (1.55) & -19.01 (1.03) & -5.87 (0.08)  \\
		\midrule
		Original PID controller  & -36.57 (0.97) & -58.98 (0.15) & -15.67 (0.68) \\
		
		\midrule
		Expert  & -5.64 (0.76) & -8.43 (0.11) & -2.31 (0.04) \\ \bottomrule
	\end{tabular}
\end{table}

\begin{figure}[]
	\centering{\includegraphics[scale=.34]{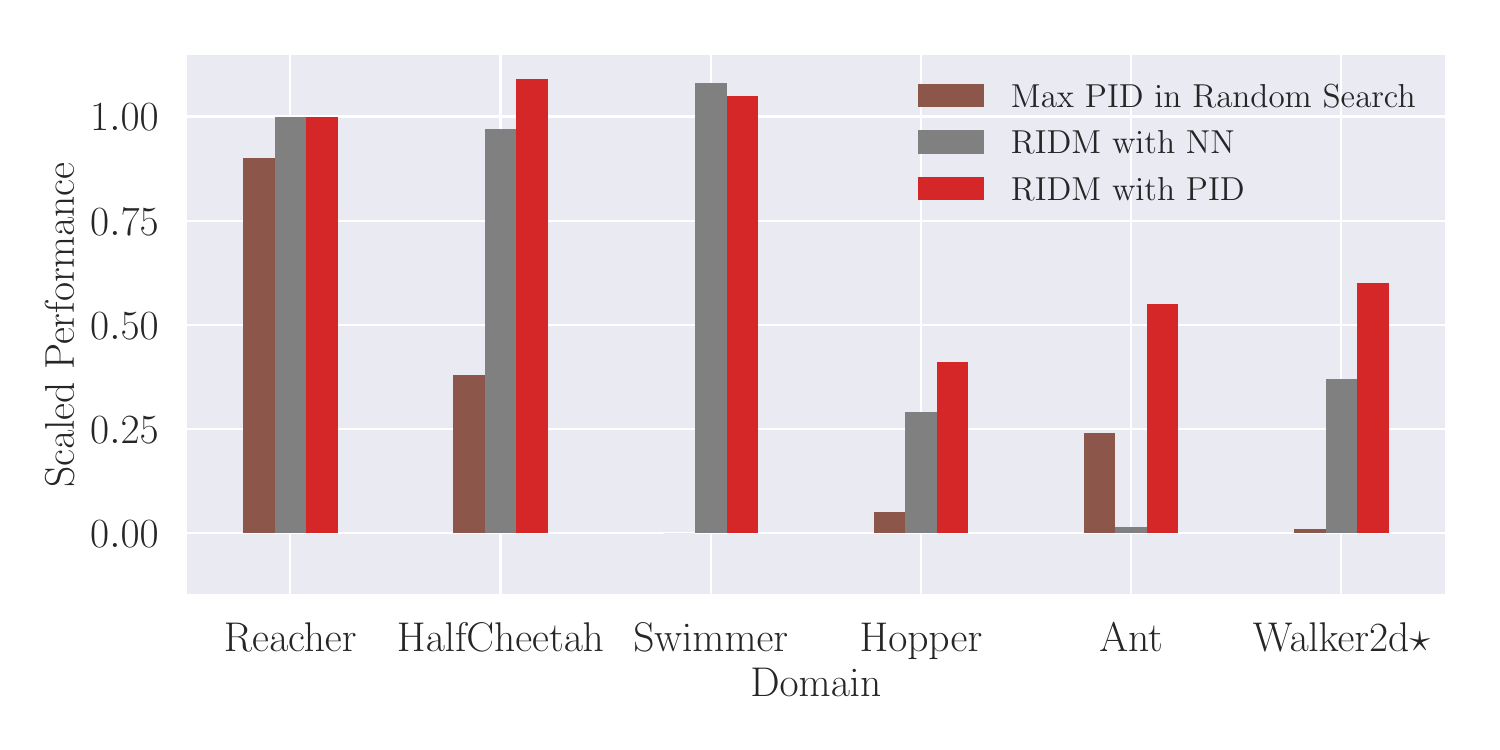}}
	\vspace{-8mm}
	\caption{Comparison of RIDM with PID as the IDM versus RIDM with NN as the IDM and the maximum performance between the randomly generated PID values on the MuJoCo domain on the same single state-only demonstration
		consisting of only raw states (exclusively of joint angles). Since RIDM with PID uses a deterministic inverse dynamics model, we do not report mean or standard deviations of our algorithm.
		$\star$PID version of RIDM used global PID gains for Walker2d-v2, unlike on other domains	where it used local PD gains.}
	\label{fig:mujoco_pid}
	\vspace{-3mm}
\end{figure}

\section{Additional Results}
\label{add_results}
Due to the high performance of RIDM with PID controllers, we performed another set of experiments on the MuJoCo domains that used a PID controller as the IDM instead of a neural
network. We compare RIDM with PID controller to 1) RIDM with NN
(from Figure \ref{fig:mujoco_nn})
and 2) best-performing randomly generated PID gains. 
We determined the best-performing random PID gain by
sampling $100$ sets of PID gains from a Gaussian distribution with mean $[0.45, 0.75, 0.15]$ and standard deviation $[.5, .5, .5]$ (each index in the list corresponds to P, I, and D),
and selecting the best-performing set. Figure \ref{fig:mujoco_pid} provides experimental results for all six of the domains.  One can see that RIDM with a PID controller performs
similarly to RIDM with an NN, and in more complex domains
such as Ant and Walker2d, significantly outperforms it. The
reasonably good performance of random PID gains shows us that
even an un-trained PID controller is an effective IDM.
RIDM with a PID controller is able to focus on optimizing just the (very few) parameters of the PID controller (i.e., the gains) as opposed to a neural network policy, where the policy space is much larger.


\section{Conclusion}
\label{discussion}

In this paper, we investigated whether or not several restrictive assumptions common to many techniques that integrate imitation and reinforcement learning -- access to demonstrator action information, access to several demonstrations, and knowledge of task-specific state augmentations -- are necessary.
We hypothesized that they are not, and we proposed a new algorithm called RIDM in order to validate that hypothesis.
RIDM is a fundamentally new method for integrated imitation and reinforcement learning that operates in scenarios for which only a single, raw-state-only demonstration is provided.
We experimentally demonstrated that RIDM can find behaviors that achieve good task performance in these scenarios. Moreover, our results show that it outperforms a reasonable baseline technique while doing so.
We posit that the success of RIDM is due to the way in which it generates behavior trajectories and performs learning -- RIDM generates behavior by directly using the demonstration data as the set points for a parameterized but robust inverse dynamics model, and iteratively optimizes the model parameters in response to the environment reward.
The above procedure not only generates reasonable trajectories over which to learn, but also reduces the learning problem to one over a relatively low-dimensional set of parameters when compared to other approaches.

This paper opens up many possible directions for future work.  For one, it may be possible to extend RIDM to learn a generalized controller from numerous demonstrations of a specific task.  For
example, in an arm-reaching task, we may have two
different demonstrations with two different target
reaching points. Another open question is how RIDM will perform when using optimization algorithms other than CMA-ES, such as TRPO or PPO. Furthermore, in all of our experiments, the state spaces are low-level features (only joint angles). Another possible future direction is to investigate how RIDM performs with video demonstrations.

\bibliographystyle{IEEEtran}
\bibliography{IEEEabrv,root}

\newpage
\newpage
\onecolumn
\section{Supplementary Materials}
\label{supp_experiments}

Here we include details of our inverse dynamics model and experiment details. 


\subsection{Proportional\textendash Integral\textendash Derivative (PID)   
	Controller}
\label{supp_pid}

The PID controller is a popular control loop feedback mechanism
used in control systems. Given that we are trying
to adjust some variable, the PID controller will help in 
accurately applying the necessary correction to reach a desired 
setpoint. For example, if we want a robot to move its arm from 
$\ang{10}$ to $\ang{30}$ (desired setpoint), the PID controller 
will appropriately calculate the necessary torque/force to 
accomplish this transition. Moreover, the PID controller is
also responsive; in other words, if the force applied to move
from $\ang{10}$ to $\ang{30}$ is less or more than required, it
will accordingly respond and adapt.

Mathematically, the PID controller is modeled as follows:

\begin{equation}
\label{pid_eq}
\begin{aligned}
u(t) &= K_p e(t) + K_i \int_0^t e(t')dt' + K_d \frac{d e(t)}{dt}
\end{aligned}
\end{equation}

where $e(t)$ is the error between the desired setpoint and current point
value, $K_p$, $K_i$, and $K_d$ are the proportionality constants for
the proportional, integral, and derivative terms respectively. Intuitively,
each term means the following: the proportional term signifies that if the desired setpoint is far from our
current point, we should apply a larger correction to reach there, the integral 
term keeps track of the cumulative error of the point from the desired
setpoint at each time step, this helps in applying a large correction if we 
have been far from the desired set point for a long time, and finally,
the derivative term represents a damping factor that controls the excessive
correction that may result from the proportional and integral components.

Since the PID controller accounts for the error to get from
one state, $s_t$, to a desired setpoint, $s_{t+1}$,
we view the PID controller as an inverse dynamics model, a mapping
from  state-transitions to actions i.e. $\{(s_t, s_{t+1})\rightarrow a_t\}$,
which tells us which action $a_t$ the agent took to go from state $s_t$
to state $s_{t+1}$. We consider input and output of Equation ~\ref{pid_eq} to be the raw states and low-level actions 
respectively.

\subsection{Experiment Details}
\label{supp_domain_dets}

\subsubsection{MuJoCo Experiments}
We train the experts for each of these domains using trust region policy
optimization (TRPO) 
\cite{DBLP:journals/corr/SchulmanLMJA15} and  proximal policy optimization (PPO)
\cite{DBLP:journals/corr/SchulmanWDRK17}, and select those
with the best performance. 
We use the hyperparameters specified in \cite{DBLP:journals/corr/SchulmanLMJA15} and \cite{DBLP:journals/corr/abs-1709-06560}. 
In our case, TRPO worked best for Reacher, HalfCheetah, 
Swimmer, and Hopper and PPO worked best for Ant and Walker2d. Details of the considered domains are as following:

\begin{itemize}
	\item Reacher. The goal is to
	move a $2$D robot arm to a fixed location. We use
	a $2$ dimensional state and action space.
	The original state space is $11$ dimensions.
	Since we simplify the state space to only joint angles, we fix 
	the target location. The reward per time-step is 
	given by
	the distance of the arm from the target per time-step and regularization
	factor of the actions.
	\item HalfCheetah. The goal is to make a cheetah
	walk as fast as possible. We use a $6$ dimensional
	state and action space. The original state space is $17$ dimensions. The reward per time-step is given by the cheetah's forward velocity
	and regularization of its actions.
	\item Swimmer. The goal is to make a snake-like
	creature swim as fast as possible in a viscous
	liquid. We use a $2$ dimensional state
	and action space. The original state space is $8$ dimensions.
	The reward per time-step is given by the swimmer's forward velocity
	and regularization of its actions.
	\item Hopper. The goal is to make a $2$D one-legged robot hop as fast as possible. We use a $3$ dimensional state
	and action space. The original state space is $11$ dimensions.
	The reward per time-step is given by the change
	in the global position of the hopper, its jump height, its forward velocity, 
	regularization of its actions, and
	its survival.
	\item Ant. The goal is to make a $4$-legged ant
	walk as fast as possible. We use an $8$ dimensional
	state and action space. The original state space is $111$ dimensions. The reward per time-step is given by the change
	in the global position of the ant, its forward velocity, 
	regularization of its actions, its contact with the surface, and
	its survival.
	\item Walker2d. The goal is to make a $2$D bipedal robot walk as fast as possible. We use a $6$ dimensional state
	and action space. The original state space is $17$ dimensions.
	The reward per time-step is given by the change
	in the global position of the walker, its walk height, its forward velocity, 
	regularization of its actions, and
	its survival.
\end{itemize}

\subsubsection{3D Simulation}
The RoboCup 3D simulation domain is supported by two components - SimSpark~\cite{boedecker2008simspark,Xu2014} and Open
Dynamics Engine (ODE). SimSpark provides support for simulated physical multiagent system
research. The ODE library enables realistic simulation of rigid body
dynamics. 

In our experiments, we are interested in imitating two tasks: (1)
speed walking and (2) long distance kick-offs. Since SimSpark does not have built-in reward 
functions, we design our own reward function. Refer to Appendix ~\ref{supp_experiments} for details about the tasks and the designed reward functions.

Since SimSpark does not have built-in reward 
functions, we design our own reward function. We note that the demonstrators may have not used
our reward function. 
\begin{itemize}
	\item Speed walking. The goal of this task is to have
	the agent walk as fast as possible while maintaining stability
	throughout the episode. To do so, we define the total reward 
	at the end of the episode to be the cumulative
	distance travelled per time-step with a $-5$
	penalty for falling down. The distance is measured
	in meters. 
	
	\item Long-distance kick-off. The goal
	of the task is to kick the ball as far as possible
	towards the center of the goal. To do so, we define the reward
	function to be 
	$$R_{kick} = (1 + x_{total}) \cdot \exp {\Big(\frac{-\theta^2}{180}\Big)} + x_{air} \cdot 100$$
	with a $-5$ penalty for slightly bumping the ball, $-10$
	penalty for falling down,
	where $x_{total}$ is the distance travelled by the ball
	along the $x$-axis, $\theta$ is the angle of
	deviation of the ball's trajectory from the
	straight line between the agent and center of
	the goal, and $x_{air}$ is the distance along the
	$x$-axis for which the ball was travelling in
	the air. $x_{total}$ and $x_{air}$ are in meters, and
	$\theta$ is in degrees. The reward function values kicks that
	travel in the air for a long distance and
	exponentially decays the reward for off-target
	kicks.
\end{itemize}

\begin{figure*}[t!]
	\centering
	\subfigure[Reacher]{\label{fig:reacher}\includegraphics[scale=0.28]{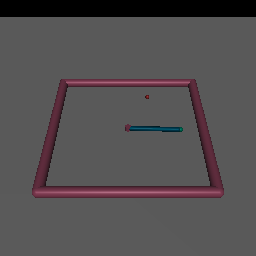}}
	\subfigure[HalfCheetah]{\label{fig:halfcheetah}\includegraphics[scale=0.28]{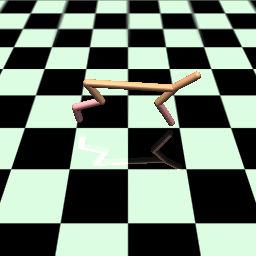}}
	\subfigure[Swimmer]{\label{fig:swimmer}\includegraphics[scale=0.28]{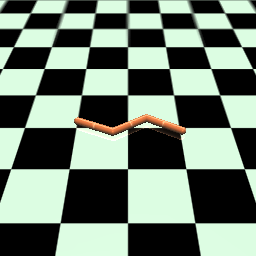}}
	\subfigure[Hopper]{\label{fig:hopper}\includegraphics[scale=0.28]{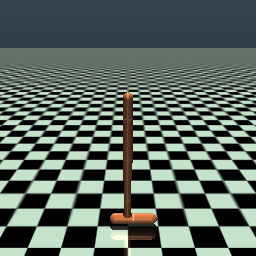}}
	\subfigure[Ant]{\label{fig:ant}\includegraphics[scale=0.28]{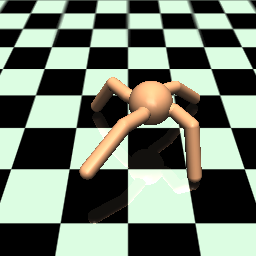}}
	%
	\subfigure[Walker2d]{\label{fig:walker}\includegraphics[scale=0.145]{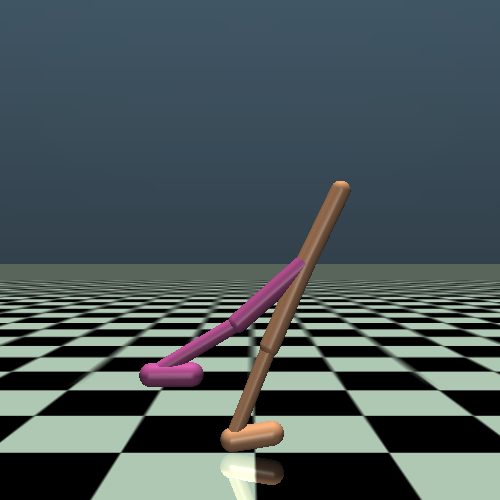}}
	\caption{Representative screenshots of the MuJoCo domains considered in this paper.}
	\label{fig:mujoco}
\end{figure*}

\begin{figure}[]
	\centering{\includegraphics[scale=0.6]{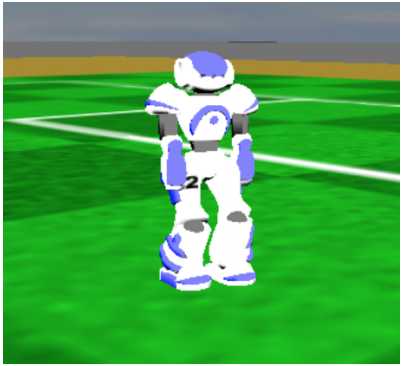}}
	\caption{Simulated Nao robot in SimSpark}
	\label{fig:nao_rob}
\end{figure}

\subsubsection{UR5 Robot Arm}

The original PID controller gains are hard-coded in the UR5 drivers when 
position-control mode is activated. Moreover, robots already include such 
a controller which is why the proposed method is so attractive, i.e., it can leverage common, pre-existing, and well-understood robotics control mechanisms.

\begin{figure}[]
	\begin{centering}
		\centering{\includegraphics[scale=0.15]{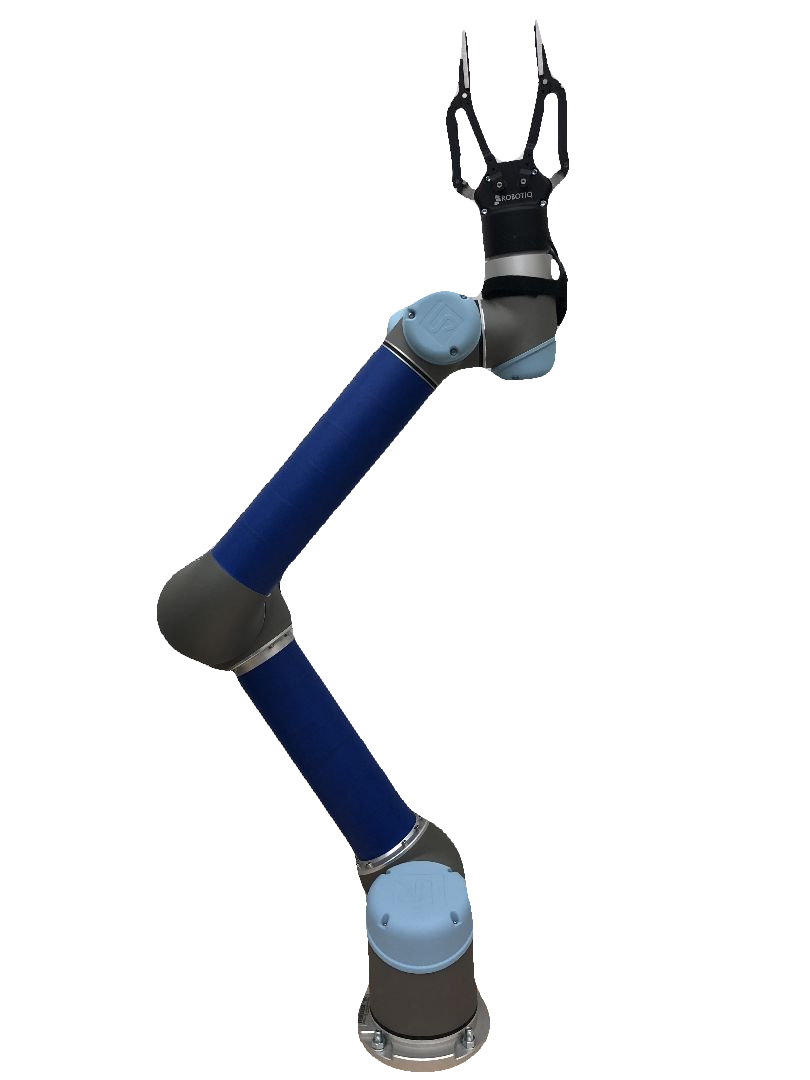}}
		\caption{UR5 Robot Arm}
		\label{fig:UR5}
	\end{centering}
\end{figure}

\end{document}